%% file: main.tex
\documentclass[runningheads]{llncs}

\usepackage{graphicx}
\usepackage{amsmath,amssymb} 
\usepackage{color}\usepackage[width=122mm,left=12mm,paperwidth=146mm,height=193mm,top=12mm,paperheight=217mm]{geometry}
\input{macros}

\newcommand{\argmin}[1]{\underset{#1}{\operatorname{arg}\,\operatorname{min}}\;}

\definecolor{darkgreen}{rgb}{0.25,0.65,0.3}
\definecolor{darkred}{rgb}{0.85,0.25,0.3}
\definecolor{darkblue}{rgb}{0.25,0.2,0.45}
\usepackage[pagebackref=true,breaklinks=true,colorlinks,bookmarks=false,citecolor=darkgreen,linkcolor=darkred,urlcolor=darkblue]{hyperref}

\begin{document}
\pagestyle{headings}
\mainmatter
\title{Precomputed Real-Time Texture Synthesis with Markovian Generative Adversarial Networks} 


\authorrunning{Chuan Li and Michael Wand}

\author{Chuan Li and Michael Wand}
\institute{Institut for Informatik, University of Mainz, Germany}

\maketitle

\begin{abstract}
This paper proposes Markovian Generative Adversarial Networks (MGANs), a method for training generative neural networks for efficient texture synthesis. While deep neural network approaches have recently demonstrated remarkable results in terms of synthesis quality, they still come at considerable computational costs (minutes of run-time for low-res images). Our paper addresses this efficiency issue. Instead of a numerical deconvolution in previous work, we precompute a feed-forward, strided convolutional network that captures the feature statistics of \emph{Markovian patches} and is able to directly generate outputs of arbitrary dimensions. Such network can directly decode brown noise to realistic texture, or photos to artistic paintings. With adversarial training, we obtain quality comparable to recent neural texture synthesis methods. As no optimization is required any longer at generation time, our run-time performance (0.25M pixel images at 25Hz) surpasses previous neural texture synthesizers by a significant margin (at least 500 times faster). We apply this idea to texture synthesis, style transfer, and video stylization.

\keywords{Texture synthesis, Adversarial Generative Networks}
\end{abstract}

\section{Introduction}

Image synthesis is a classical problem in computer graphics and vision~\cite{Efros01,Wei00}. The key challenges are to capture the structure of complex classes of images in a concise, learnable model, and to find efficient algorithms for learning such models and synthesizing new image data. Most traditional ``\emph{texture synthesis}'' methods address the complexity constraints using Markov random field (MRF) models that characterize images by statistics of local patches of pixels. 

Recently, generative models based on deep neural networks have shown exciting new perspectives for image synthesis~\cite{Goodfellow2014,Gatys15,GregorDGW15}. Deep architectures capture appearance variations in object classes beyond the abilities of pixel-level approaches. However, there are still strong limitations of how much structure can be learned from limited training data. This currently leaves us with two main classes of ``deep'' generative models: 1) \emph{full-image models} that generate whole images~\cite{Goodfellow2014,Denton15}, and 2) \emph{Markovian models} that also synthesize textures~\cite{Gatys15,LiW16}.

The first class, full-image models, are often designed as specially trained auto-encoders~\cite{KingmaW13,GregorDGW15}. Results are impressive but limited to rather small images (typically around 64$\times$64 pixels) with limited fidelity in details. The second class, the deep Markovian models, capture the statistics of local patches only and assemble them to high-resolution images. Consequently, the fidelity of details is good, but additional guidance is required if non-trivial global structure should be reproduced~\cite{Efros01,Hertzmann01,Barnes09,Gatys15,LiW16}. Our paper addresses this second approach of deep Markovian texture synthesis.

Previous neural methods of this type~\cite{Gatys15,LiW16} are built upon a deconvolutional framework~\cite{Zeiler14,Mordvintsev15}. This naturally provides blending of patches and permits reusing the intricate, emergent multi-level feature representations of large, discriminatively trained neural networks like the VGG network~\cite{Simonyan14c}, repurposing them for image synthesis. As a side note, we will later observe that this is actually crucial for high-quality result (Figure~\ref{fig:fig_MDANGeneralization}). Gatys et al.~\cite{Gatys15} pioneer this approach by modeling patch statistics with a global Gaussian models of the higher-level feature vectors, and Li et al.~\cite{LiW16} utilize dictionaries of extended local patches of neural activation, trading-off flexibility for visual realism.

Deep Markovian models are able to produce remarkable visual results, far beyond traditional pixel-level MRF methods. Unfortunately, the run-time costs of the deconvolution approach are still very high, requiring iterative back-propagation in order to estimate a pre-image (pixels) of the feature activations (higher network layer). In the most expensive case of modeling MRFs of higher-level feature patches~\cite{LiW16}, a high-end GPU needs several minutes to synthesize low-resolution images (such as a 512-by-512 pixels image).

The objective of our paper is therefore to improve the efficiency of deep Markovian texture synthesis. The key idea is to precompute the inversion of the network by fitting a strided\footnote{A strided convolutional network hence replaces pooling layers by subsampled convolution filters that learn pooling during training (for example, two-fold mean pooling is equivalent to blurred convolution kernels sampled at half resolution).} convolutional network~\cite{Springenberg2015,RadfordMC15} to the inversion process, which operates purely in a feed-forward fashion. Despite being trained on patches of a fixed size, the resulting network can generate continuous images of arbitrary dimension without any additional optimization or blending, yielding a high-quality texture synthesizer of a specific style and high performance\footnote{See supplementary material and code at: https://github.com/chuanli11/MGANs}.

We train the convolutional network using adversarial training~\cite{RadfordMC15}, which permits maintaining image quality similar to the original, expensive optimization approach. As result, we obtain significant speed-up: Our GPU implementation computes $512 \times 512$ images within 40ms (on an nVidia TitanX). The key limitation, of course, is to precompute the feed-forward convolutional network for each texture style. Nonetheless, this is still an attractive trade-off for many potential applications, for example from the area of artistic image or video stylization. We explore some of these applications in our experiments. 
 
\section{Related Work}
Deconvolutional neural networks have been introduced to visualize deep features and object classes. Zeiler et al.~\cite{Zeiler14} back-project neural activations to pixels. Mahendran et al.~\cite{Mahendran15} reconstruct images from the neural encoding in intermediate layers. Recently, effort are made to improve the efficiency and accuracy of deep visualization~\cite{YosinskiCNFL15,NguyenYC16}. Mordvintsev et al. have raised wide attention by showing how deconvolution of class-specifc activations can create hallucinogenic imagery from discriminative networks~\cite{Mordvintsev15}. The astonishing complexity of the obtained visual patterns has immediately spurred hope for new generative models: Gatys et al.~\cite{Gatys15,Gatys2015b} drove deconvolution by global covariance statistics of feature vectors on higher network layers, obtaining unprecedented results in artistic style transfer. The statistical model has some limitations: Enforcing per-feature-vector statistics permits a mixing of feature patterns that never appear in actual images and limit plausibility of the learned texture. This can be partially addressed by replacing point-wise feature statistics by statistics of spatial patches of feature activations~\cite{LiW16}. This permits photo-realistic synthesis in some cases, but also reduces invariance because the simplistic dictionary of patches introduces rigidity. On the theory side, Xie et al.~\cite{Xie16} have proved that a generative random field model can be derived from used discriminative networks, and show applications to unguided texture synthesis.
 
Full image methods employ specially trained auto-encoders as generative networks~\cite{KingmaW13}. For example, the Generative Adversarial Networks (GANs) use two networks, one as the discriminator and other as the generator, to iteratively improve the model by playing a minimax game~\cite{Goodfellow2014}. This model is extended to work with a Laplacian pyramid~\cite{Gauthier15}, and with a conditional setting~\cite{Denton15}. Very recently, Radford et al.~\cite{RadfordMC15} propose a set of architectural refinements\footnote{strided convolution, ReLUs, batch normalization, removing fully connected layers} that stabilized the performance of this model, and show that the generators have vector arithmetic properties. One important strength of adversarial networks is that it offers perceptual metrics~\cite{LarsenSW15,DosovitskiyB16} that allows auto-encoders to be training more efficiently. These models can also be augmented semantic attributes~\cite{YanYSL15}, image captions~\cite{MansimovPBS15}, 3D data~\cite{DosovitskiySB14,KulkarniWKT15}, spatial/temporal status~\cite{GregorDGW15,ImKJM16,OordKK16} etc.

In very recent, two concurrent work, Ulyanov et al.~\cite{Ulyanov16} and Johnson et al.~\cite{Johnson16} propose fast implementations of Gatys et al's approach. Both of their methods employ precomputed decoders trained with a perceptual texture loss and obtain significant run-time benefits (higher decoder complexity reduces their speed-up a bit). The main conceptual difference in our paper is the use of Li et al.'s~\cite{LiW16} feature-patch statistics as opposed to learning Gaussian distributions of individual feature vectors, which provides some benefits in terms of reproducing textures more faithfully. 

\section{Model}
\begin{figure*}[t]
	\centering
		\vspace{-1mm} \includegraphics[width=0.85\linewidth]{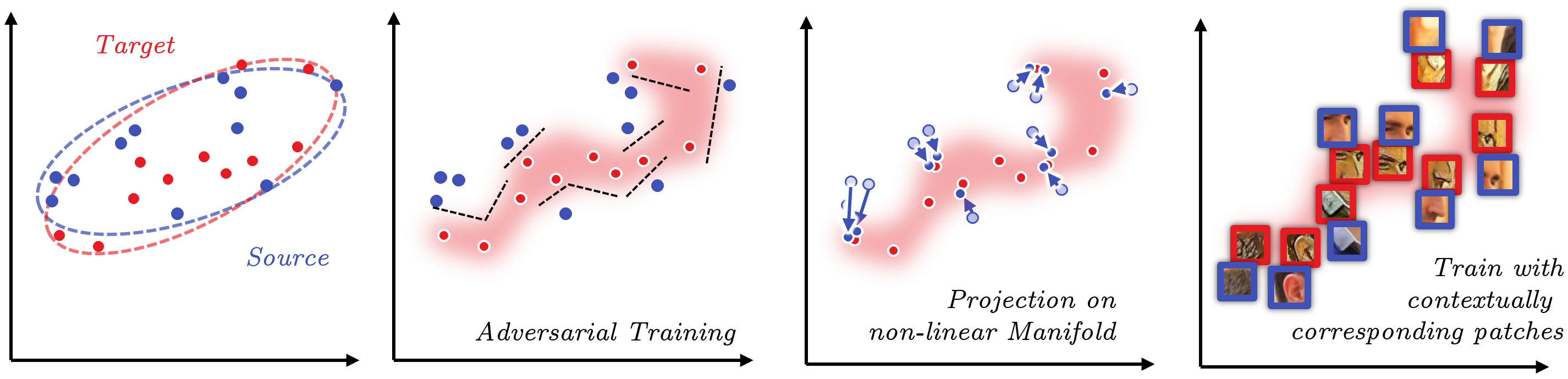} \vspace{-1mm}
		\caption{Motivation: real world data does not always comply with a Gaussian distribution (first), but a complex nonlinear manifold (second). We adversarially learn a mapping to project contextually related patches to that manifold.}
	\label{fig:Model}
	\vspace*{-2mm}
\end{figure*}

Let us first conceptually motive our method. Statistics based methods~\cite{Gatys15,Ulyanov16} match the distributions of source (input photo or noise signal) and target (texture) with a Gaussian model (Figure~\ref{fig:Model}, first). They do not further improve the result once two distributions match. However, real world data does not always comply with a Gaussian distribution. For example, it can follow a complicated non-linear manifold. Adversarial training~\cite{Goodfellow2014} can recognize such manifold with its discriminative network (Figure~\ref{fig:Model}, second), and strengthen its generative power with a projection on the manifold (Figure~\ref{fig:Model}, third). We improve adversarial training with contextually corresponding Markovian patches (Figure~\ref{fig:Model}, fourth). This allows the learning to focus on the mapping between different depictions of the same context, rather than the mixture of context and depictions.

Figure~\ref{fig:Architecture} visualizes our pipeline, which extends the patch-based synthesis algorithm of Li et al.~\cite{LiW16}. We first replace their patch dictionary (including the iterative nearest-neighbor search) with a continuous discriminative network \textit{D} (green blocks) that learns to distinguish actual feature patches (on VGG\_19 layer Relu3\_1, purple block) from inappropriately synthesized ones. A second comparison (pipeline below \textit{D}) with a VGG\_19 encoding of the same image on the higher, more abstract layer Relu5\_1 can be optionally used for guidance. If we run deconvolution on the VGG networks (from the discriminator and optionally from the guidance content), we obtain deconvolutional image synthesizer, which we call \emph{Markovian Deconvolutional Adversarial Networks} (MDANs).

MDANs are still very slow; therefore, we aim for an additional generative network \textit{G} (blue blocks; a strided convolutional network). It takes a VGG\_19 layer Relu4\_1 encoding of an image and directly decodes it to pixels of the synthesis image. During all of the training we do not change the \textit{VGG\_19} network (gray blocks), and only optimize \textit{D} and \textit{G}. Importantly, both \textit{D} and \textit{G} are trained simultaneously to maximize the quality of \textit{G}; \textit{D} acts here as adversary to \textit{G}. We denote the overall architecture by \emph{Markovian Generative Adversarial Networks} (MGANs).

\subsection{Markovian Deconvolutional Adversarial Networks (MDANs)}
\begin{figure*}[t]
	\centering
		\includegraphics[width=0.8\linewidth]{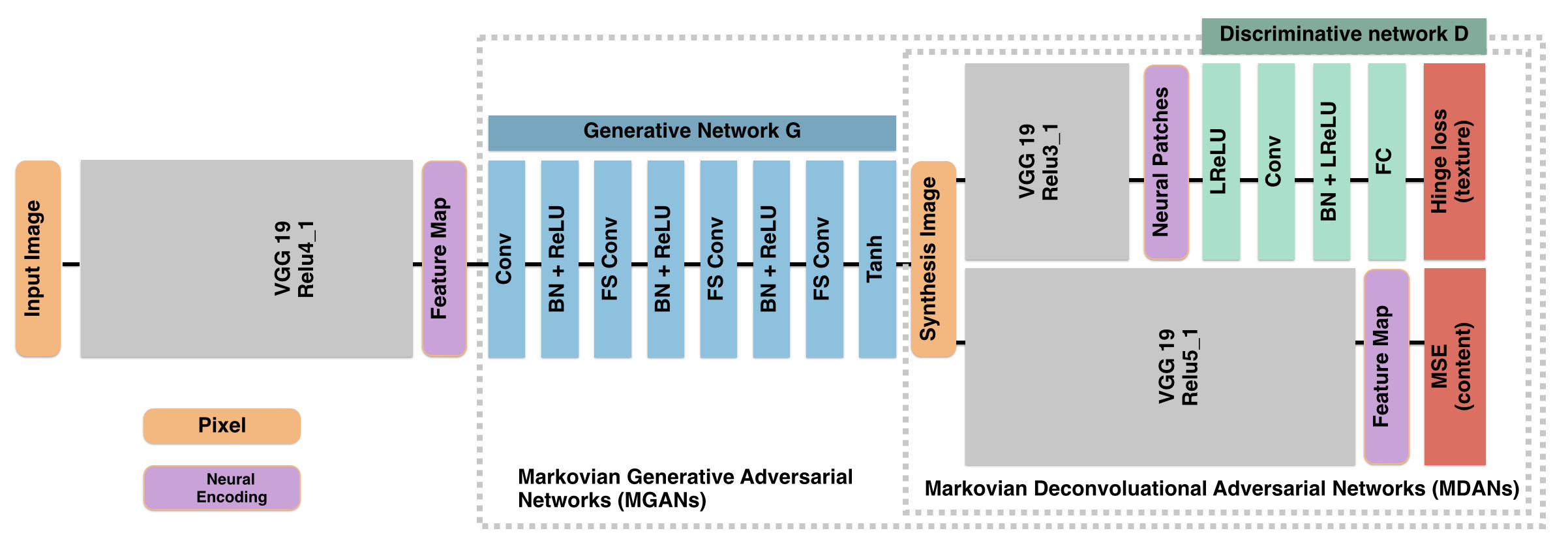}
		\caption{Our model contains a generative network (blue blocks) and a discriminative network (green blocks). We apply the discriminative training on Markovian neural patches (purple block as the input of the discriminative network.).}
	\label{fig:Architecture}
	\vspace*{-2mm}
\end{figure*}

Our MDANs synthesize textures with a deconvolutional process that is driven by adversarial training: a discriminative network \textit{D} (green blocks in Figure~\ref{fig:Architecture}) is trained to distinguish between ``neural patches'' sampled from the synthesis image and sampled from the example image. We use regular sampling on layer \textit{relu3\_1} of \textit{VGG\_19} output (purple block). It outputs a classification score $s= \pm1$ for each neural patch, indicating how ``real'' the patch is (with $s = 1$ being real). For each patch sampled from the synthesized image, $1 - s$ is its texture loss to minimize. The deconvolution process back-propagates this loss to pixels. Like Radford et al.~\cite{RadfordMC15} we use batch normalization (BN) and leaky ReLU (LReLU) to improve the training of \textit{D}. 

Formally, we denote the example texture image by $\bv{x}_t \in \reals^{w_t \times h_t}$, and the synthesized image by $\bv{x} \in \reals^{w \times h}$. We initialize $\bv{x}$ with random noise for un-guided synthesis, or an content image $\bv{x}_c \in \reals^{w \times h}$ for guided synthesis. The deconvolutio iteratively updates $\bv{x}$ so the following energy is minimized: 
\begin{eqnarray}
\bv{x}&= &\argmin{x}E_{t}(\Phi(\bv{x}), \Phi(\bv{x}_{t})) + \alpha_{1}E_{c}(\Phi(\bv{x}), \Phi(\bv{x}_{c})) + \alpha_{2}\Upsilon(\bv{x})
\label{eq:MDAN}
\end{eqnarray}
Here $E_{t}$ denotes the texture loss, in which $\Phi(\bv{x})$ is $\bv{x}$'s feature map output from layer \textit{relu3\_1} of \textit{VGG\_19}. We sample patches from $\Phi(\bv{x})$, and compute $E_{t}$ as the Hinge loss with their labels fixed to one: 
\begin{equation}
E_{t}(\Phi(\bv{x}), \Phi(\bv{x}_{t})) = \frac{1}{N}\sum_{i = 1}^{N}\max(0, 1 - 1 \times s_{i})
\label{eq:hingeloss}
\end{equation}
Here $s_i$ denotes the classification score of $i$-th neural patch, and $N$ is the total number of sampled patches in $\Phi(\bv{x})$. The discriminative network is trained on the fly: Its parameters are randomly initialized, and then updated after each deconvolution, so it becomes increasingly smarter as synthesis results improve. 

\begin{figure*}[t]
	\centering
		\includegraphics[width=0.8\linewidth]{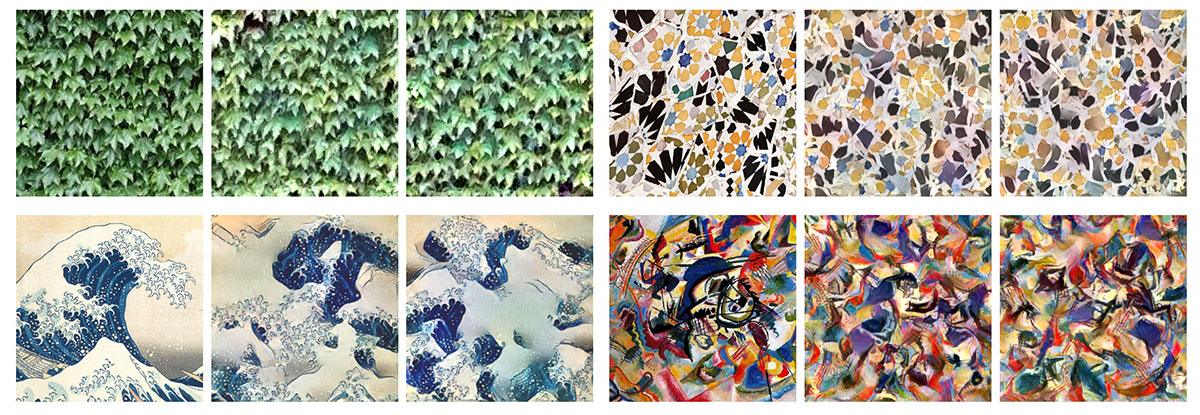}
		\caption{Un-guided texture synthesis using MDANs. For each case the first image is the example texture, and the other two are the synthesis results. Image credits: ~\cite{Xie16}'s ``Ivy'', flickr user erwin brevis's ``güell'', Katsushika Hokusai's ``The Great Wave off Kanagawa'', Kandinsky's ``Composition VII''.}
	\label{fig:MDAN_results_UnGuided}
\end{figure*}

\begin{figure*}[t]
	\centering
		\includegraphics[width=0.8\linewidth]{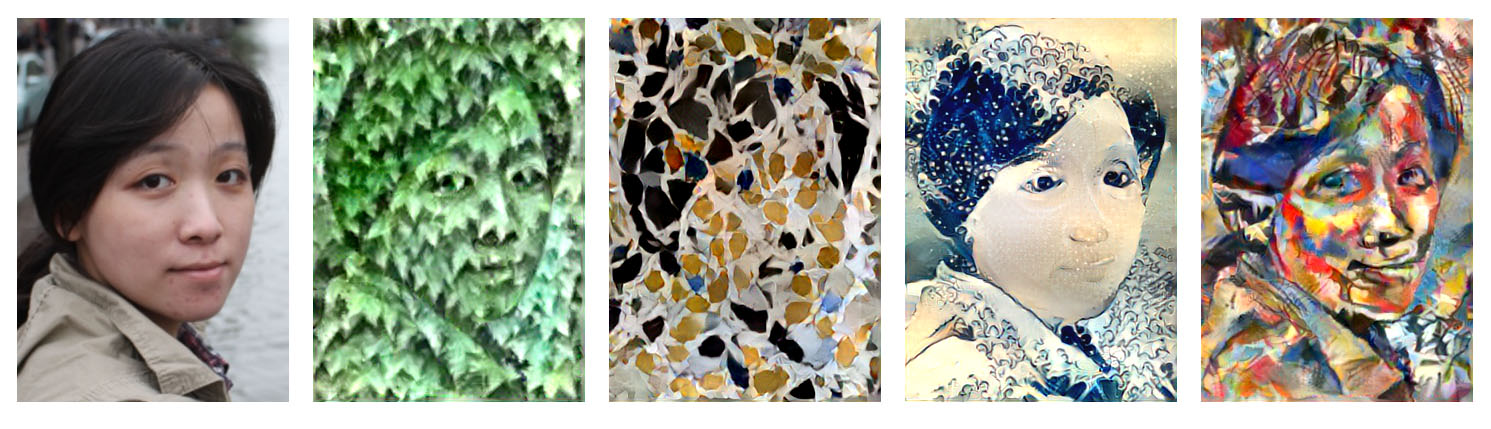}
		\caption{Guided texture synthesis using MDANs. The reference textures are the same as in Figure~\ref{fig:MDAN_results_UnGuided}.}
	\label{fig:MDAN_results_Guided}
	\vspace*{-5mm}
\end{figure*}

The additional regularizer $\Upsilon(\bv{x})$ in Eq.~\ref{eq:MDAN} is a smoothness prior for pixels~\cite{Mahendran15}. Using $E_{t}$ and $\Upsilon(\bv{x})$ can synthesize random textures (Figure~\ref{fig:MDAN_results_UnGuided}). By minimizing an additional content loss $E_{c}$, the network can generate an image that is contextually related to a guidance image $\bv{x}_{c}$(Figure~\ref{fig:MDAN_results_Guided}). This content loss is the Mean Squared Error between two feature maps $\Phi(\bv{x})$ and $\Phi(\bv{x}_{c})$. We set the weights with $\alpha_1 = 1$ and $\alpha_2 = 0.0001$, and minimize Equation~\ref{eq:MDAN} using back-propagation with ADAM~\cite{KingmaB14} (learning rate 0.02, momentum 0.5). Notice each neural patch receives its own output gradient through the back-propagation of \textit{D}. In order to have a coherent transition between adjacent patches, we blend their output gradient like texture optimization~\cite{Kwatra05} did. 

\subsection{Markovian Generative Adversarial Networks (MGANs)}
MDANs require many iterations and a separate run for each output image. We now train a variational auto-encoder (VAE) that decodes a feature map directly to pixels. The target examples (textured photos) are obtained from the MDANs. Our generator \textit{G} (blue blocks in Figure~\ref{fig:Architecture}) takes the layer \textit{relu4\_1} of \textit{VGG\_19} as the input, and decodes a picture through a ordinary convolution followed by a cascade of fractional-strided convolutions (FS Conv). Although being trained with fixed size input, the generator naturally extends to arbitrary size images.

\begin{figure*}[t]
	\centering
		\includegraphics[width=0.9\linewidth]{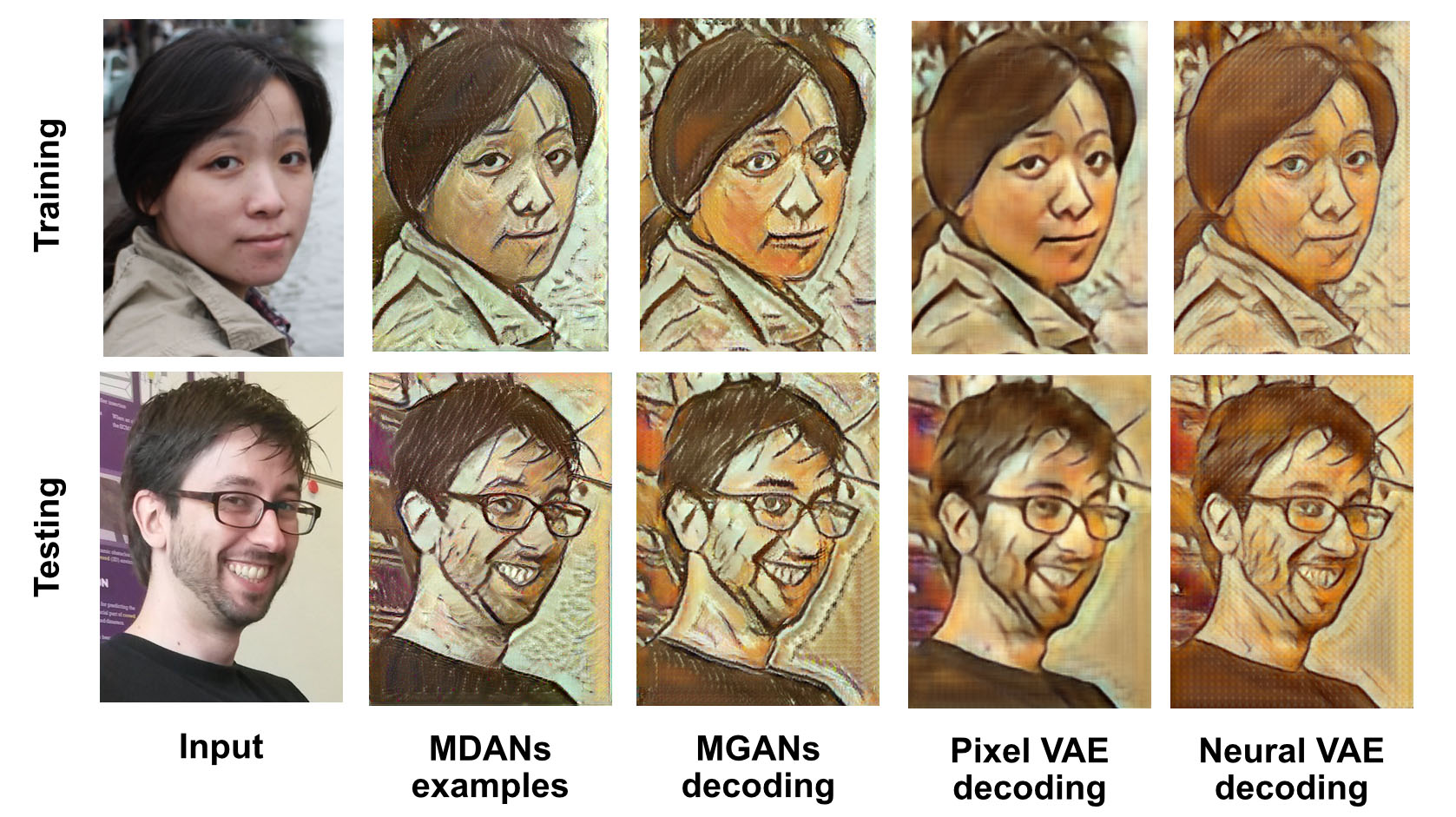}
		\caption{Our MGANs learn a mapping from \textit{VGG\_19} encoding of the input photo to the stylized example (MDANs). The reference style texture for MDANs is Pablo Picasso's ``self portrait 1907''. We compare the results of MGANs to Pixel VAE and Neural VAE in with both training and testing data.}
	\label{fig:MGAN_results}
	\vspace*{-5mm}
\end{figure*}

As Dosovitskiy et al.~\cite{DosovitskiyB16} point out, it is crucially important to find a good metric for training an auto-encoder: Using the Euclidean distance between the synthesized image and the target image at the pixel level (Figure~\ref{fig:MGAN_results}, pixel VAE) yields an over-smoothed image. Comparing at the neural encoding level improves results (Figure~\ref{fig:MGAN_results}, neural VAE), and adversarial training improves the reproduction of the intended style further (Figure~\ref{fig:MGAN_results}, MGANs).

Our approach is similar to classical Generative Adversarial Networks (GANs) \cite{Goodfellow2014}, with the key difference of not operating on full images, but neural patches from the \emph{same} image. Doing so utilizes the contextual correspondence between the patches, and makes learning easier and more effective in contrast to learning the distribution of a object class~\cite{Goodfellow2014} or a mapping between contextually irrelevant data~\cite{Ulyanov16}. In additional we also replace the Sigmoid function and the binary cross entropy criteria from~\cite{RadfordMC15} by a max margin criteria (Hinge loss). This avoids the vanishing gradient problem when learning \textit{D}. This is more problematic in our case than in Radfort et al.'s~\cite{RadfordMC15} because of less diversity in our training data. Thus, the Sigmoid function can be easily saturated. 

\begin{figure*}[t]
	\centering
		\includegraphics[width=0.8\linewidth]{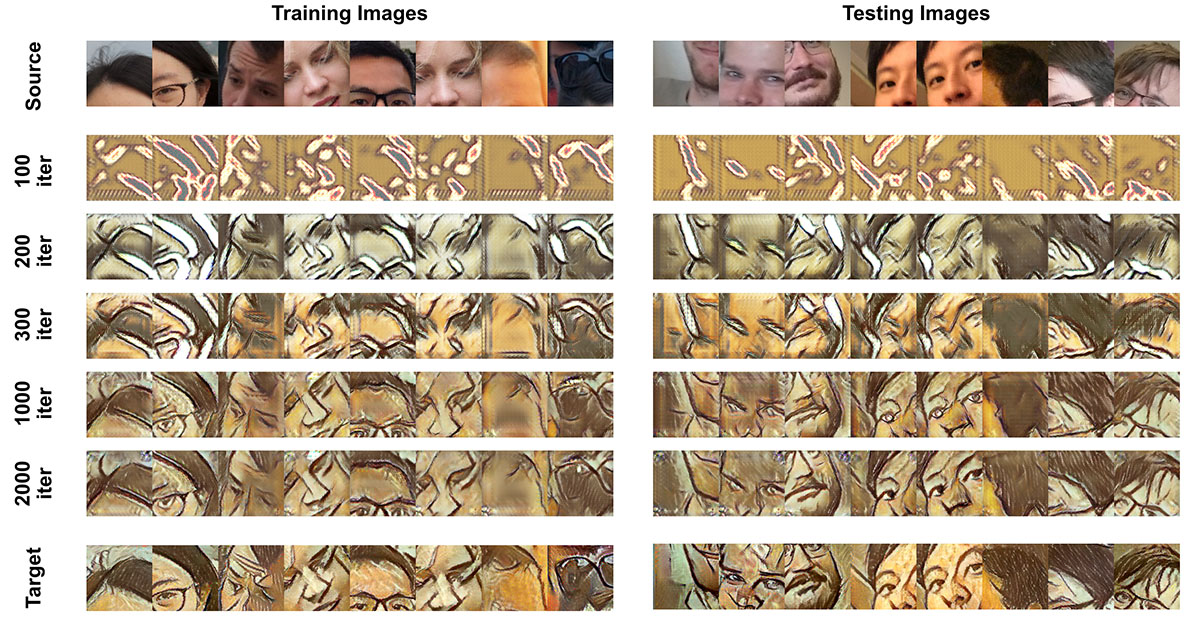}
		\caption{Intermediate decoding results during the training of MGANs. The reference style texture for MDANs is Pablo Picasso's ``self portrait 1907''.}
	\label{fig:MGAN_TrainingProcess}
\end{figure*}

Figure~\ref{fig:MGAN_results} (MGANs) shows the results of a network that is trained to produce paintings in the style of Picasso's ``Self-portrait 1907''. For training, we randomly selected 75 faces photos from the CelebA data set~\cite{liu2015celebA}, and in additional to it 25 non-celebrity photos from the public domain. We resize all photos so that the maximum dimension is 384 pixels. We augmented the training data by generating 9 copies of each photo with different rotations and scales. We regularly sample subwindows of 128-by-128 croppings from them for batch processing. In total we have 24,506 training examples, each is treated as a training image where neural patches are sampled from its \textit{relu3\_1} encoding as the input of \textit{D}.

Figure~\ref{fig:MGAN_results} (top row, MGANs) shows the decoding result of our generative network for a training photo. The bottom row shows the network generalizes well to test data. Notice the MDANs image for the test image is never used in the training. Nonetheless, direct decoding with \textit{G} produces very good approximation of it. The main difference between MDANs and MGANs is: MDANs preserve the content of the input photo better and MGANs produce results that are more stylized. This is because MGANs was trained with many images, hence learned the most frequent features. Another noticeable difference is MDANs create more natural backgrounds (such as regions with flat color), due to its iterative refinement. Despite such flaws, the MGANs model produces comparable results with a speed that is 25,000 times faster. 

Figure~\ref{fig:MGAN_TrainingProcess} shows some intermediate results MGANs. It is clear that the decoder gets better with more training. After 100 batches, the network is able to learn the overall color, and where the regions of strong contrast are. After 300 batches the network started to produce textures for brush strokes. After 1000 batches it learns how to paint eyes. Further training is able to remove some of the ghosting artifacts in the results. Notice the model generalizes well to testing data (right).

\section{Experimental Analysis}
\begin{figure*}[t]
	\centering
		\includegraphics[width=0.9\linewidth]{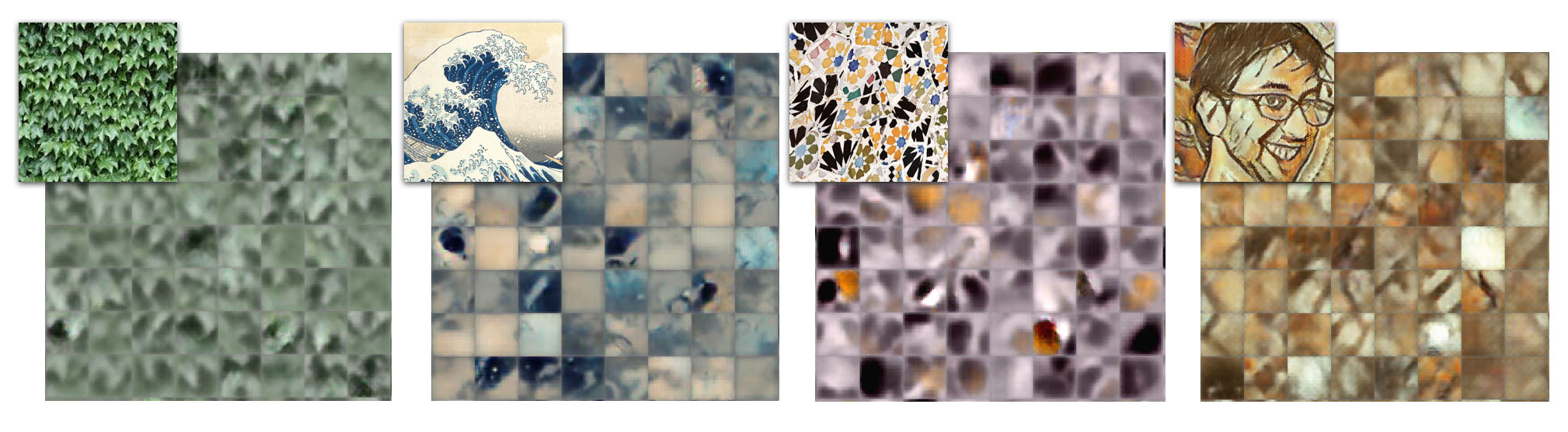}
		\caption{Visualizing the learned features in the generative networks. Image credits:~\cite{Xie16}'s ``Ivy'', flickr user erwin brevis's ``güell'', Katsushika Hokusai's ``The Great Wave off Kanagawa''.}
	\label{fig:analysis_vis_filters}
\end{figure*}

We conduct empirical experiments with our model: we study parameter influence (layers for classification, patch size) and the complexity of the model (number of layers in the network, number of channels in each layer). While there may not be a universal optimal design for all textures, our study shed some light on how the model behaves for different cases. For fair comparison, we scale the example textures in this study to fixed size (128-by-128 pixels), and demand the synthesis output to be 256-by-256 pixels.

\textbf{Visualizing decoder features:} We visualize the learned filters of decoder \textit{G} in Figure~\ref{fig:analysis_vis_filters}. These features are directly decoded from a one-hot input vector. Individual patches are similar to, but not very faithfully matching the example textures (reconfirming the semi-distributed and non-linear nature of the encoding). Nonetheless, visual similarity of such artificial responses seems strong enough for synthesizing new images.

\begin{figure*}[t]
	\centering
		\includegraphics[width=0.9\linewidth]{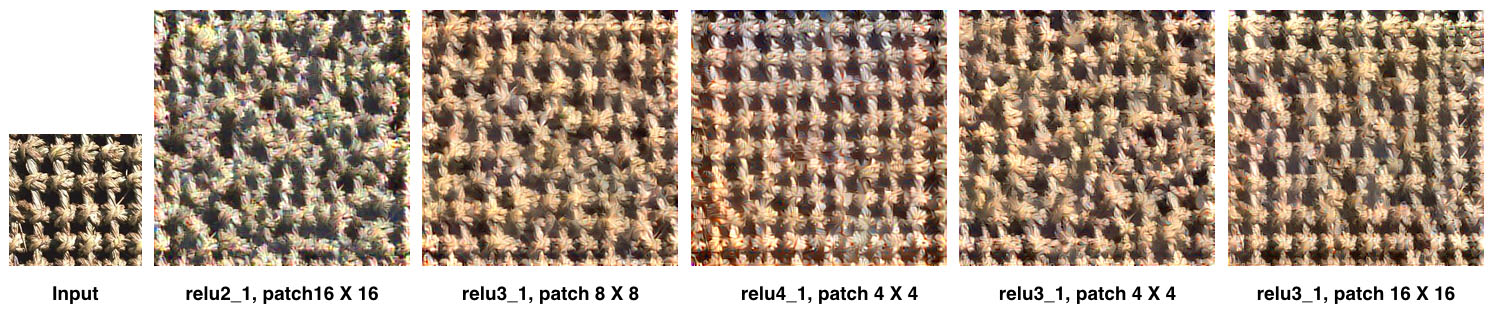}
		\caption{Different layers and patch sizes for training the discriminative netowrk. Input image credit: ``ropenet'' from the project link of~\cite{Kwatra2003GC}.}
	\label{fig:analysis_MDAN_differentlayers}
	\vspace*{-5mm}
\end{figure*}

\textbf{Parameters:} Next, we study the influence of changing the input layers for the discriminative network. To do so we run unguided texture synthesis with discriminator \textit{D} taking layer \textit{relu2\_1}, \textit{relu3\_1}, and \textit{relu4\_1} of \textit{VGG\_19} as the input. We use patch sizes of 16, 8 and 4 respectively for the three options, so they have the same receptive field of 32 image pixels (approximately; ignoring padding). The first three results in Fig.~\ref{fig:analysis_MDAN_differentlayers} shows the results of these three settings. Lower layers (\textit{relu2\_1}) produce sharper appearances but at the cost of losing form and structure of the texture. Higher layer (\textit{relu4\_1}) preserves coarse structure better (such as regularity) but at the risk of being too rigid for guided scenarios. Layer \textit{relu3\_1} offers a good balance between quality and flexibility. We then show the influence of patch size: We fix the input layer of \textit{D} to be \textit{relu3\_1}, and compare patch size of 4 and 16 to with the default setting of 8. The last two results in Fig.~\ref{fig:analysis_MDAN_differentlayers} shows that such change will also affect the rigidity of the model: smaller patches increase the flexibility and larger patches preserve better structure. 

\begin{figure*}[t]
	\centering
		\includegraphics[width=0.9\linewidth]{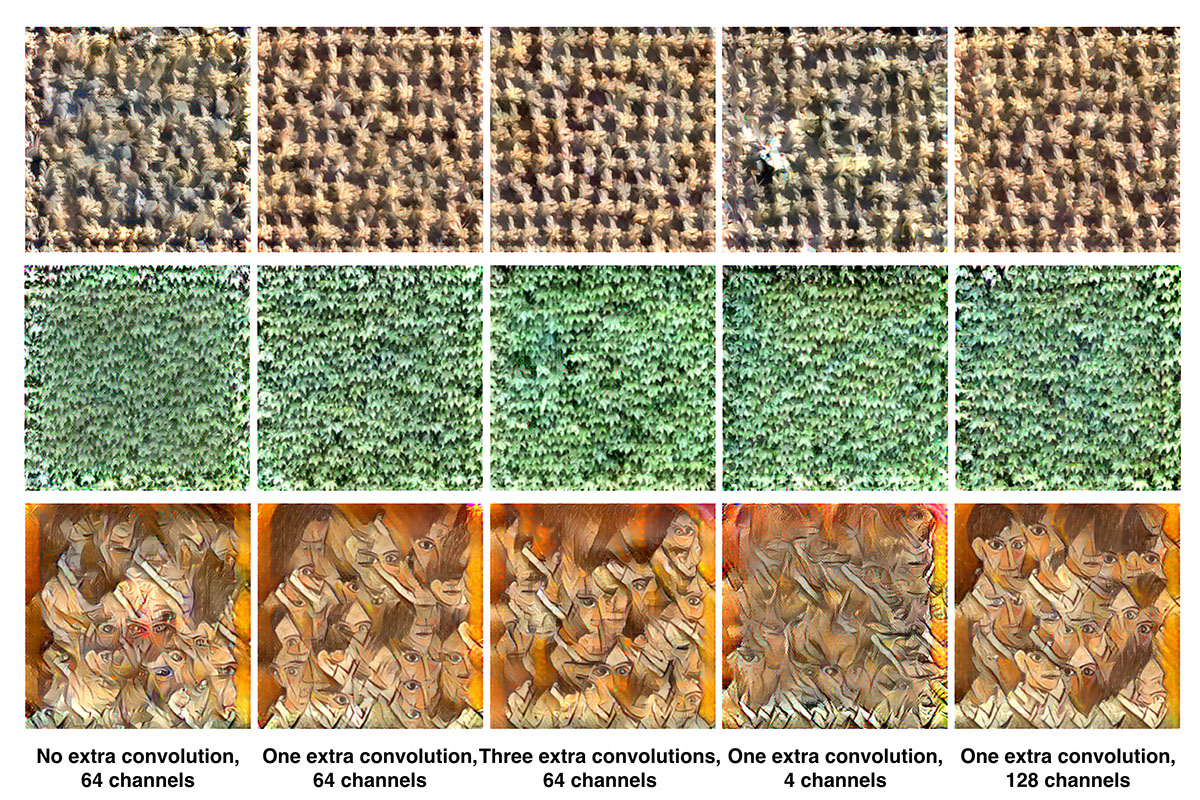}
		\caption{Different depths for training the discriminative netowrk. The input textures are ``ropenet'' from the project link of~\cite{Kwatra2003GC},~\cite{Xie16}'s ``Ivy'', and Pablo Picasso's ``self portrait 1907''.}
	\label{fig:analysis_Depth_netD}
	\vspace*{-5mm}
\end{figure*}

\textbf{Complexity:} We now study the influence of 1) the number of layers in the networks and 2) the number of channels in each layer. We first vary the \textit{D} by removing the convolutional layer. Doing so reduces the depth of the network and in consequence the synthesis quality (first column, Fig.~\ref{fig:analysis_Depth_netD}). Bringing this convolutional layer back produces smoother synthesis (second column, Fig.~\ref{fig:analysis_Depth_netD}). However, in these examples the quality does not obviously improves with more additional layers (third column, Fig.~\ref{fig:analysis_Depth_netD}).

Testing the \textit{D} with 4, 64, and 128 channels for the convolutional layer, we observe in general that decreasing the number of channels leads to worse results (fourth column, Fig.~\ref{fig:analysis_Depth_netD}), but there is no significance difference between 64 channels and 128 channels (second column v.s. fifth column). The complexity requirements also depend on the actual texture. For example, the ivy texture is a rather simple MRF, so the difference between 4 channel and 64 channel are marginal, unlike in the other two cases.

Next, we fix the discriminative network and vary the complexity of the generative network. We notice some quality loss when removing the first convolutional layer from the decoder, or reducing the number of channels for all layers, and only very limited improvement from a more complex design. However the difference is not very significant. This is likely because the networks are all driven by the same discriminative network, and the reluctance of further improvement indicates there are some non-trivial information from the deconvolutional process that can not be recovered by a feed forward process. In particular, the fractionally-strided convolutions does not model the nonlinear behaviour of the max-pooling layer, hence often produces alias patterns. These become visible in homogeneous, texture-less area. To avoid artifacts but encourage texture variability, we can optionally add Perlin noise~\cite{Perlin85} to the input image.

\subsubsection{Initialization} 
\begin{figure*}[t]
	\centering
		\includegraphics[width=0.9\linewidth]{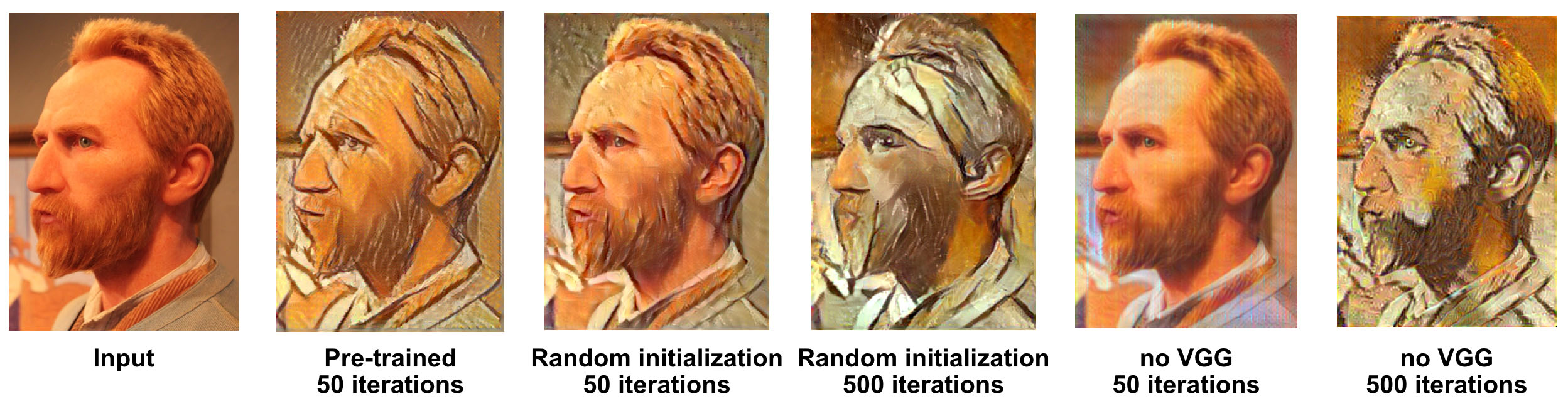}
		\caption{Different initializations of the discriminative networks. The reference texture is Pablo Picasso's ``self portrait 1907''.}
	\label{fig:fig_MDANGeneralization}
\end{figure*}

Usually, networks are initialized with random values. However we found \textit{D} has certain generalization ability. Thus, for transferring the same texture to different images with MDANs, a previously trained network can serve as initialization. Figure~\ref{fig:fig_MDANGeneralization} shows initialization with pre-trained discriminative network (that has already transferred 50 face images) produces good result with only 50 iterations. In comparison, random initialization does not produce comparable quality even after the first 500 iterations. It is useful to initialize \textit{G} with an auto-encoder that directly decodes the input feature to the original input photo. Doing so essentially approximates the process of inverting \textit{VGG\_19}, and let the whole adversarial network to be trained more stably.

\textbf{The role of VGG:} We also validate the importance of the pre-trained \textit{VGG\_19} network. As the last two pictures in Figure~\ref{fig:fig_MDANGeneralization} show, training a discriminative network from scratch (from pixel to class label~\cite{RadfordMC15}) yields significantly worse results. This has also been observed by Ulyanov et al.~\cite{Ulyanov16}. Our explanation is that much of the statistical power of VGG\_19 stems from building shared feature cascades for a diverse set of images, thereby approaching human visual perception more closely than a network trained with a limited example set.

\section{Results}
This section shows examples of our MGANs synthesis results. We train each model with 100 randomly selected images from ImageNet, and a single example texture. We first produce 100 transferred images using the MDANs model, then regularly sample 128-by-128 image croppings as training data for MGANs. In total we have around 16k samples for each model. The training take as about 12 min per epoch. Each epoch min-batches through all samples in random order. We train each texture for upto five epochs. 

\begin{figure*}[t]
	\centering
		\includegraphics[width=0.9\linewidth]{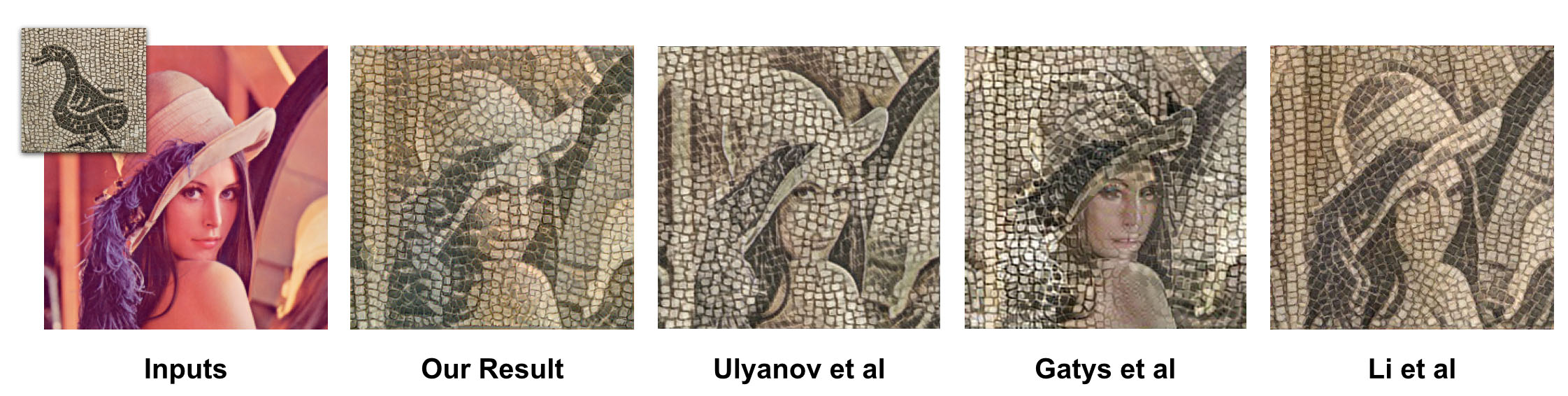}
		\caption{Comparisons with previous methods. See more examples in our supplementary report. Results of Ulyanov et al.~\cite{Ulyanov16}, Gatys et al.~\cite{Gatys15} and input images are from~\cite{Ulyanov16}.}
	\label{fig:Gallery_CompareALL}
\end{figure*}

Figure~\ref{fig:Gallery_CompareALL} compares our synthesis results with previous methods. First, our method has a very different character in comparison to the methods that use global statistics~\cite{Ulyanov16,Gatys15}: It transfers texture more coherently, such as the hair of Lena was consistently mapped to dark textures. In contrast, the Gaussian model ~\cite{Ulyanov16,Gatys15} failed to keep such consistency, and have difficulty in transferring complicated image content. For example the eyes in ~\cite{Ulyanov16}'s result and the entire face in~\cite{Gatys15}'s result are not textured. Since these features do not fit a Gaussian distribution, they are difficult to be constrained by a Gram matrix. The other local patch based approach~\cite{LiW16} produces the most coherent synthesis, due to the use of non-parametric sampling. However, their method requires patch matching so is significantly slower (generate this 384-by-384 picture in 110 seconds). Our method and Ulyanov et al.~\cite{Ulyanov16} run at the same level of speed; both bring significantly improvement of speed over Gatys et al.~\cite{Gatys15} (500 times faster) and Li et al.~\cite{LiW16} (5000 times faster). 


\begin{figure*}[t]
	\centering
		\includegraphics[width=1\linewidth]{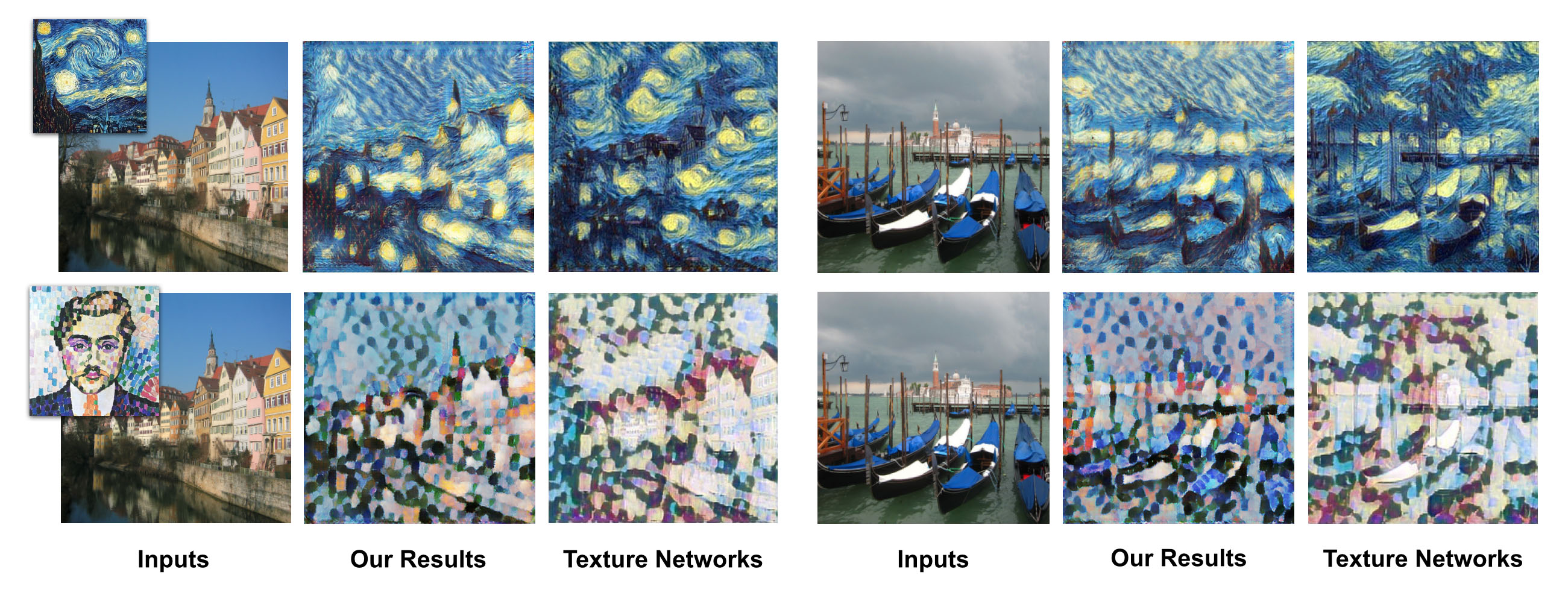}
		\caption{More comparisons with Texture Networks~\cite{Ulyanov16}. Results of~\cite{Ulyanov16} and input images are from~\cite{Ulyanov16}.}
	\label{fig:Gallery_TextureNet}
\end{figure*}

Figure~\ref{fig:Gallery_TextureNet} further discuss the difference between the Gaussian based method~\cite{Ulyanov16} and our method\footnote{Since Ulyanov et al.~\cite{Ulyanov16} and Johnson et al.~\cite{Johnson16} are very similar approaches, in this paper we only compare to one of them~\cite{Ulyanov16}. The main differences of~\cite{Johnson16} are: 1) using a residual architecture instead of concatenating the outputs from different layers; 2) no additional noise in the decoding process.}. In general~\cite{Ulyanov16} produces more faithful color distributions in respect to the style image. It also texture background better (see the starry night example) due to the learning of mapping from noise to Gaussian distribution. On the other hand, our method produces more coherent texture transfer and does not suffer the incapability of Gaussian model for more complex scenarios, such as the facade in both examples. In comparison~\cite{Ulyanov16} produces either too much or too little textures in such complex regions. 

\begin{figure*}[t]
	\centering
		\includegraphics[width=0.9\linewidth]{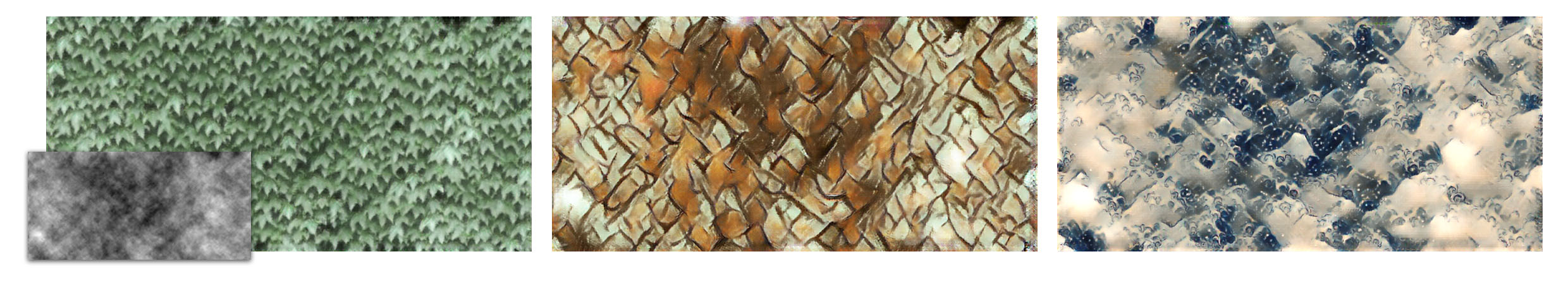}
		\caption{Generate random textures by decoding from Brown noise.}
	\label{fig:UnGuided}
\end{figure*}

Figure~\ref{fig:UnGuided} shows that unguided texture synthesis is possible by using the trained model to decode noise input. In this case, Perlin noise\footnote{We need to use ``brown'' noise with spectrum decaying to the higher frequencies because flat ``white'' noise creates an almost flat response in the encoding of the VGG network. Somer lower-frequency structure is required to trigger the feature detectors in the discriminative network.} images are forwarded through \textit{VGG\_19} to generate feature maps for the decoder. To our surprise, the model that was trained with random ImageNet images is able to decode such features maps to plausible textures. This again shows the generalization ability of our model. Last, Figure~\ref{fig:UnGuided} shows our video decoding result. As a feed-forward process our method is not only faster but also relatively more temporally coherent than the deconvolutional methods.

\begin{figure*}[t]
	\centering
		\includegraphics[width=0.9\linewidth]{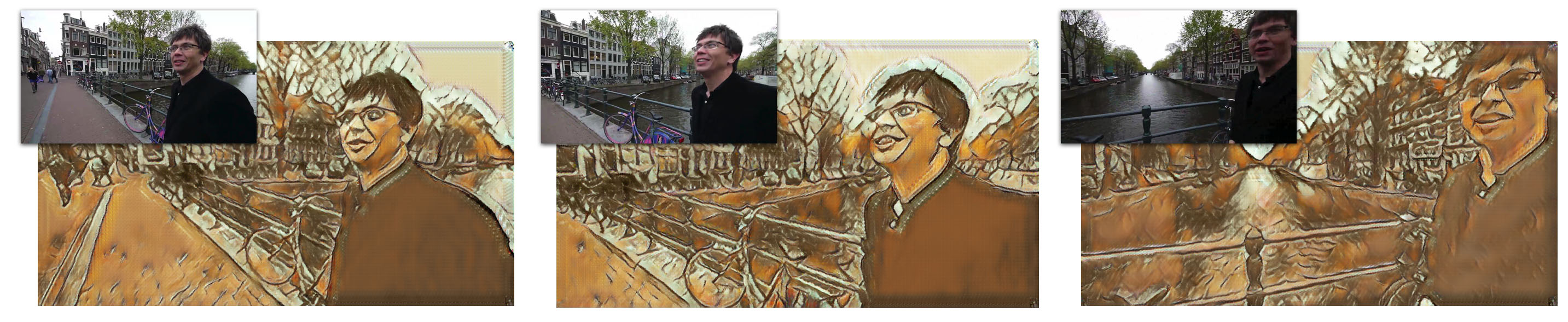}
		\caption{Decoding a 1080-by-810 video. We achieved the speed of 8Hz. Input video is credited to flickr user macro antonio torres.}
	\label{fig:Video}
\end{figure*}

\begin{figure*}[ht]
	\centering
		\includegraphics[width=1\linewidth]{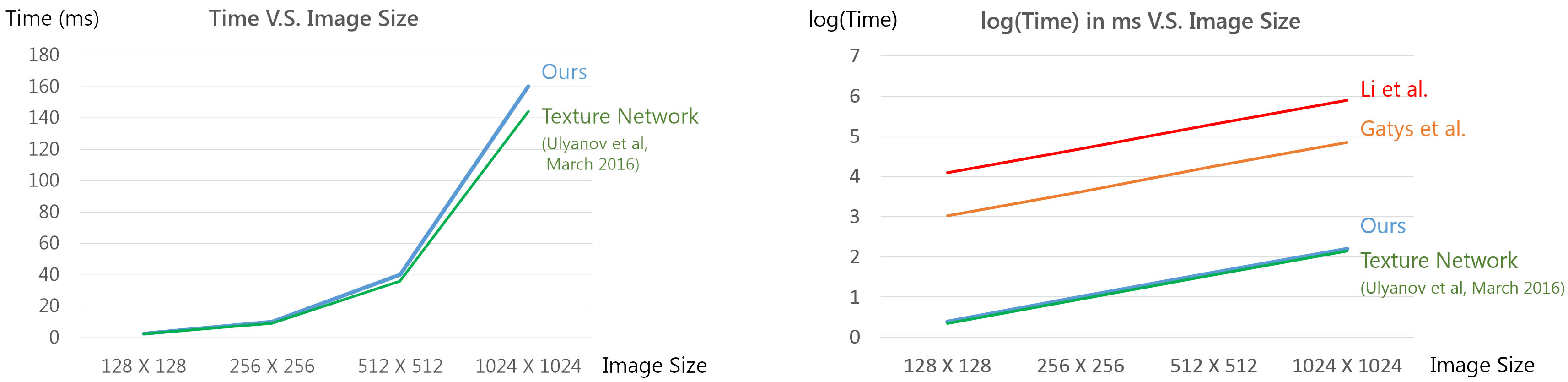}
		\caption{Left: speed comparison between our method and Ulyanov et al.~\cite{Ulyanov16}.  Right: speed comparison (in log space) between our method and Gatys et al.~\cite{Gatys15}, Li et al.~\cite{LiW16}, and Ulyanov et al.~\cite{Ulyanov16}. The feed-forward methods (ours and~\cite{Ulyanov16}) are significantly faster than Gatys et al.~\cite{Gatys15} (500 times speed up) and Li et al.~\cite{LiW16} (5000 times speed up).}
	\label{fig:Time}
\end{figure*}

Last but not the least, we provide details for the time/memory usage of our method. The time measurement is based on a standard benchmark framework~\cite{SoumithBench} (Figure~\ref{fig:Time}): Our speed is at the same level as the concurrent work by Ulyanov et al.~\cite{Ulyanov16}, who also use a feed-forward approach, perform significantly faster than previous deconvolution based approaches~\cite{Gatys15,LiW16}. More precisely, both our method and Ulyanov et al.~\cite{Ulyanov16} are able to decode 512-by-512 images at 25Hz (Figure~\ref{fig:Time}, left), while~\cite{Ulyanov16} leads the race by a very small margin. The time cost of both methods scale linearly with the number of pixels in the image. For example, our method cost 10 ms for a 256-by-256 image, 40 ms for a 512-by-512 image, and 160 ms for a 1024-by-1024 image. Both methods show a very significant improvement in speed over previous deconvolutional methods such as Gatys et al.~\cite{Gatys15} and Li et al.~\cite{LiW16} (Figure~\ref{fig:Time} right): about 500 times faster than Gatys et al.~\cite{Gatys15}, and 5000 times faster than Li et al.~\cite{LiW16}. In the meantime our method is also faster than most traditional pixel based texture synthesizers (which rely on expensive nearest-neighbor searching). A possible exceptions would be a GPU implementation of ``Patch Match''~\cite{Barnes09}, which could run at comparable speed. However, it provides the quality benefits (better blending, invariance) of a deep-neural-network method (as established in previous work~\cite{Gatys15,LiW16}).

Memory-wise, our generative model takes 70 Mb memory for its parameters(including the \textit{VGG} network till layer Relu4\_1). At runtime, the required memory to decode a image linearly depends on the image's size: for a 256-by-256 picture it takes about 600 Mb, and for a 512-by-512 picture it requires about 2.5 Gb memory. Notice memory usage can be reduced by subdividing the input photo into blocks and run the decoding in a scanline fashion. However, we do not further explore the optimization of memory usage in this paper. 

\section{Limitation}
Our current method works less well with non-texture data. For example, it failed to transfer facial features between two difference face photos. This is because facial features can not be treated as textures, and need semantic understanding (such as expression, pose, gender etc.). A possible solution is to couple our model with the learning of object class~\cite{RadfordMC15} so the local statistics is better conditioned. For synthesizing photo-realistic textures, Li et al.~\cite{LiW16} often produces better results due to its non-parametric sampling that prohibits data distortion. However, the rigidity of their model restricts its application domain. Our method works better with deformable textures, and runs significantly faster. 

Our model has a very different character compared to Gaussian based models~\cite{Gatys15,Ulyanov16}. By capturing a global feature distribution, these other methods are able to better preserve the global ``look and feels'' of the example texture. In contrast, our model may deviate from the example texture in, for example, the global color distribution. However, such deviation may not always be bad when the content image is expected to play a more important role. 

Since our model learns the mapping between different depictions of the same content, it requires features highly invariant features. For this reason we use the pre-trained \textit{VGG\_19} network. This makes our method weaker in dealing with highly stationary backgrounds (sky, out of focus region etc.) due to their weak activation from \textit{VGG\_19}. We observed that in general statistics based methods~\cite{Ulyanov16,Gatys15} generate better textures for areas that has weak content, and our method works better for areas that consist of recognizable features. We believe it is valuable future work to combine the strength of both methods.

Finally, we discuss the noticeable difference between the results of MDANs and MGANs. The output of MGANs is often more consistent with the example texture, this shows MGANs' strength of learning from big data. MGANs has weakness in flat regions due to the lack of iterative optimization. More sophisticated architectures such as the recurrent neural networks can bring in state information that may improve the result. 

\section{Conclusion}
The key insight of this paper is that adversarial generative networks can be applied in a Markovian setting to learn the mapping between different depictions of the same content. We develop a fully generative model that is trained from a single texture example and randomly selected images from ImageNet. Once trained, our model can decode brown noise to realistic texture, or photos into artworks. We show our model has certain advantages over the statistics based methods~\cite{Ulyanov16,Gatys15} in preserving coherent texture for complex image content. Once trained (which takes about an hour per example), synthesis is extremely fast and offers very attractive invariance for style transfer.

Our method is only one step in the direction of learning generative models for images. An important avenue for future work would be to study the broader framework in a big-data scenario to learn not only Markovian models but also include coarse-scale structure models. This additional invariance to image layout could, as a side effect, open up ways to also use more training data for the Markovian model, thus permitting more complex decoders with stronger generalization capability over larger classes. The ultimate goal would be a directly decoding, generative image model of large classes of real-world images.

\section*{Acknowledgments}
This work has been partially supported by the Intel Visual Computing Institute and the Center for Computational Science Mainz. We like to thank Bertil Schmidt and Christian Hundt for providing additional computational resources.

\bibliographystyle{splncs03}
\bibliography{mybib}
\end{document}

%% file: macros.tex
\usepackage{color}
\definecolor{ChrisCol}{RGB}{255,50,50}
\definecolor{MichaelCol}{RGB}{100,120,240}

\usepackage{amsmath}
\usepackage{amsfonts}
\usepackage{amssymb}
\usepackage{algorithm}
\usepackage[noend]{algpseudocode}

\newcommand{\reals}{\mathbb{R}}
\newcommand{\bv}[1]{\mathbf{#1}}

\usepackage{lineno}
\usepackage{multirow}
\usepackage{array}
\usepackage{stackrel}
\usepackage{tabularx}
\usepackage{subfig}
\usepackage{url}